\pdfoutput=1

\documentclass[11pt]{article}

\usepackage[final]{acl}

\usepackage{times}
\usepackage{latexsym}

\usepackage[T1]{fontenc}

\usepackage[utf8]{inputenc}

\usepackage{microtype}

\usepackage{inconsolata}

\usepackage{graphicx}

\usepackage{multirow}
\usepackage{verbatim}
\usepackage{multicol}
\usepackage{mathtools}
\usepackage{booktabs}
\usepackage[flushleft]{threeparttable}
\usepackage{tablefootnote}

\RequirePackage{etoolbox}
\RequirePackage{xcolor}
\definecolor{Gray}{gray}{0.9}
\definecolor{cb-black}      {RGB}{ 0,   0,   0}
\definecolor{cb-blue-green} {RGB}{ 0,  073,  073}
\definecolor{cb-green-sea}  {RGB}{ 0, 146, 146}
\definecolor{cb-rose}       {RGB}{255, 109, 182}
\definecolor{cb-salmon-pink}{RGB}{255, 182, 119}
\definecolor{cb-purple}     {RGB}{ 73,   0, 146}
\definecolor{cb-blue}       {RGB}{ 0, 109, 219}
\definecolor{cb-lilac}      {RGB}{182, 109, 255}
\definecolor{cb-blue-sky}   {RGB}{109, 182, 255}
\definecolor{cb-blue-light} {RGB}{182, 219, 255}
\definecolor{cb-burgundy}   {RGB}{146,   0,   0}
\definecolor{cb-brown}      {RGB}{146,  73,   0}
\definecolor{cb-clay}       {RGB}{219, 209,   0}
\definecolor{cb-green-lime} {RGB}{ 36, 255,  36}
\definecolor{cb-yellow}     {RGB}{255, 255, 109}

\usepackage{fontawesome5}

\newcommand{\papilusion}[0]{\raisebox{-.25\height}{\includegraphics[width=.06\textwidth]{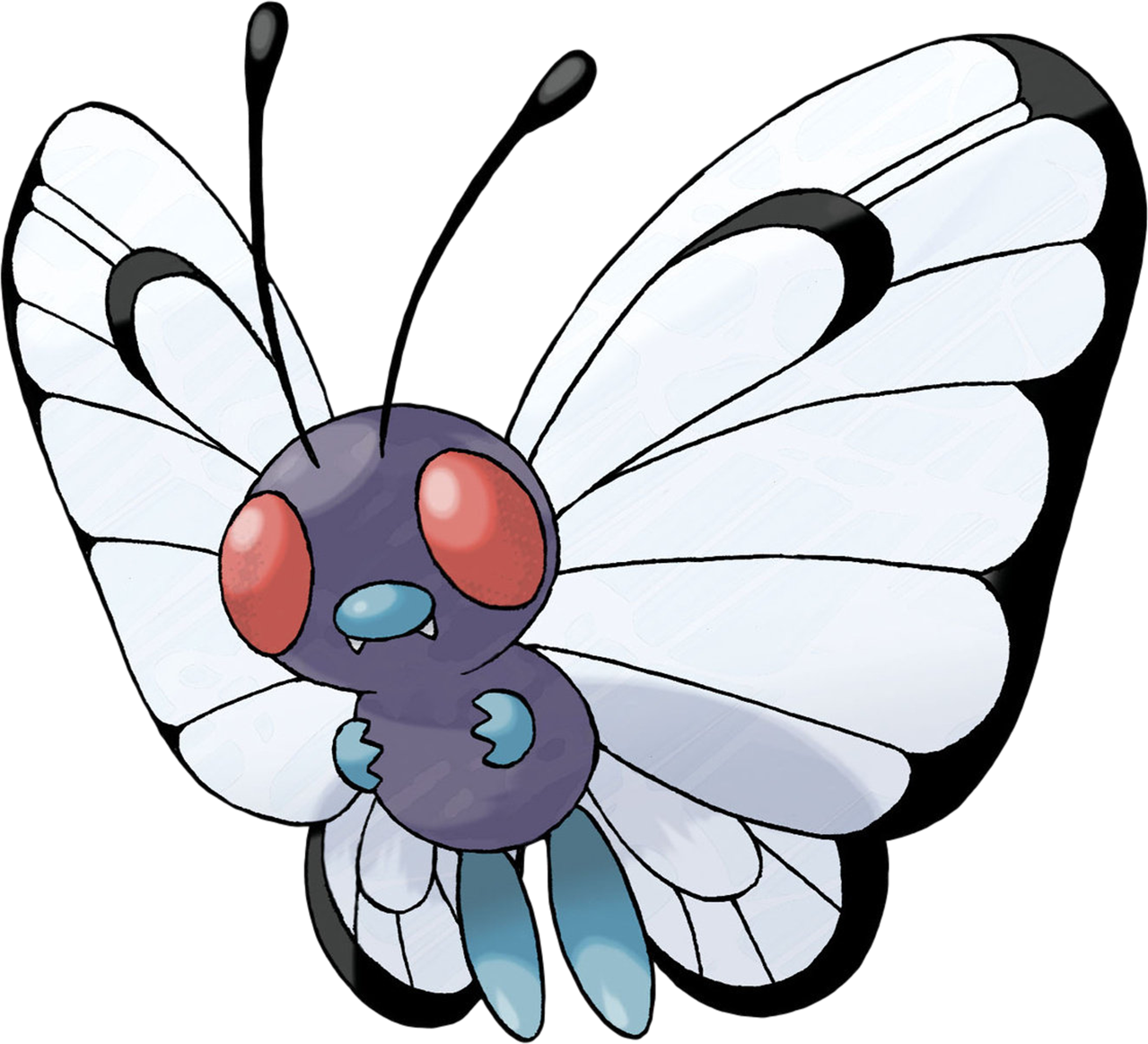}}} 

\title{\papilusion \ Papilusion at DAGPap24: Paper or Illusion? Detecting AI-generated Scientific Papers
}

\author{ Nikita Andreev\textsuperscript{\faTree}, 
        Alexander Shirnin \textsuperscript{\faCrow} \\
     \textbf{Vladislav Mikhailov}\textsuperscript{\faSnowflake},
    \textbf{Ekaterina Artemova}\textsuperscript{\faPaw} \\
    \textsuperscript{\faTree}CAIT and Applied AI Institute,
    \textsuperscript{\faCrow}HSE University \\
    \textsuperscript{\faSnowflake}University of Oslo,
    \textsuperscript{\faPaw}Toloka AI  \\
    \small{
    \textbf{Correspondence:} \href{mailto:}{\texttt{ashirnin@hse.ru}}
}
}

\begin{document}
\maketitle
\begin{abstract}
This paper presents Papilusion, an AI-generated scientific text detector developed within the DAGPap24 shared task on detecting automatically generated scientific papers. We propose an ensemble-based approach and conduct ablation studies to analyze the effect of the detector configurations on the performance. Papilusion is ranked 6th on the leaderboard, and we improve our performance after the competition ended, achieving 99.46 (+9.63) of the F1-score on the official test set.
\end{abstract}

\section{Introduction}
A series of the DAGPap shared tasks addresses mitigation of the risks associated with misusing language technologies in the scientific domain, such as fabricating research publications \cite{else2021fight,van2021hundreds} and affecting the peer review ecosystem at the leading AI conferences \cite{liang2024monitoring}. DAGPap22 \cite{kashnitsky-etal-2022-overview} focuses on a well-established task formulation in the artificial text detection field \cite{uchendu2023reverse}, which requires to determine if a given text is human-written or AI-generated. DAGPap24\footnote{\href{https://github.com/ChamezopoulosSavvas/DAGPap24}{\texttt{github.com/ChamezopoulosSavvas/DAGPap24}}} offers a less explored task formulation, which is framed as a token-level classification problem. Developing robust and reliable AI-generated scientific text detectors promotes analyzing scientific text collections at scale and assisting non-expert and expert users in identifying the fabricated content \cite{cabanac2021tortured,else2023abstracts}. 

This paper proposes Papilusion\footnote{Papilusion is named after a pokémon \textit{Papilusion}, which is the French name for \textit{Butterfree}.}, an AI-generated text detection system for tagging text segments, which are produced by a generative language model (LM) or modified through synonym replacement and summarization. The token-level AI-generated text detection accounts for a more practical scenario, where a generative LM is used as a writing assistant. Papilusion is an ensemble of encoder-only sequence taggers that leverages the strengths of multiple independently finetuned LMs to enhance the accuracy of detecting AI-generated text. We conduct ablation studies to analyze the impact of various hyperparameter and LM configurations on the target performance. 

Papilusion is ranked 6th among 30 participating teams at DAGPap24, with 89.83 of the F1-score on the leaderboard. After the competition ends, we improve our solution achieving 99.46 (+9.63) of the F1-score on the official evaluation set.

\begin{figure*}[ht!]
    \centering
    \includegraphics[width=0.7\textwidth]{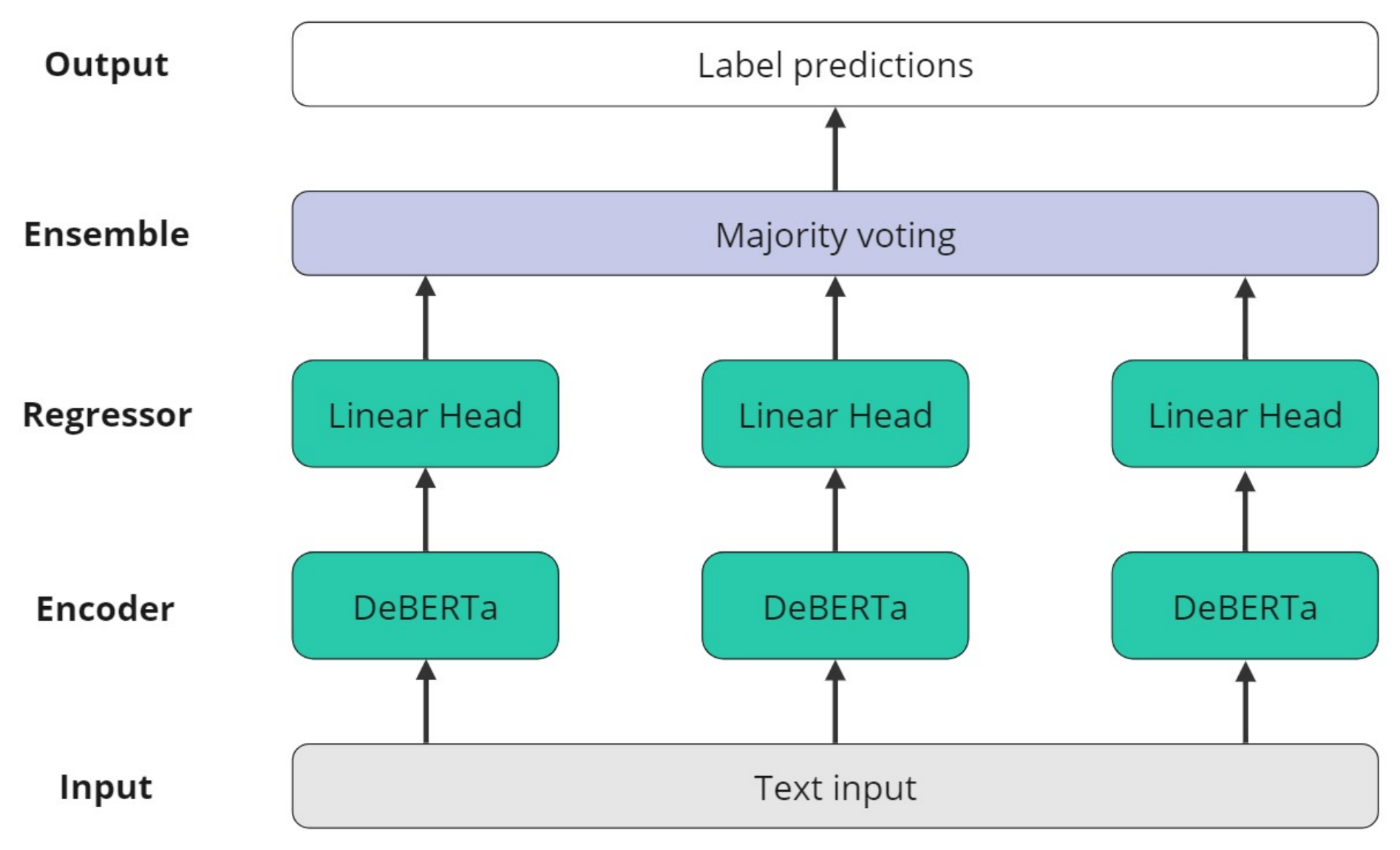}
    \caption{The Papilusion pipeline involves fine-tuning three distinct encoder models, which are based on the same architecture but trained independently with different hyperparameters. These models use linear heads to predict labels that differentiate between human-written and machine-generated text. Finally, a majority vote is applied to aggregate the predicted labels.}
    \label{fig:pipeline}
\end{figure*}

\section{Background}
The DAGPap24 dataset is a collection of human-machine mixed scientific texts, where the text segments can be (i) human-written, (ii) modified through an NLTK-based synonym replacement \cite{bird-loper-2004-nltk}, (iii) produced by ChatGPT, and (iv) summarized. The organizers provide 5000, 5000, and 20000 training, development, and test examples, respectively. The average number of tokens\footnote{The tokens are provided by the organizers and are obtained by splitting a text based on whitespace characters.} is 5591.

\paragraph{Task Formulation} The task is to assign one of the four corresponding labels to each token in a given text, as it was described above. The example for a text with a synonym replacement is shown below. Here, ``\texttt{1}'' stands for a synonym replacement, and ``\texttt{0}'' refers to the human-written segment:

\begin{itemize}
    \item \texttt{text:} ``\textcolor{cb-burgundy}{this was continued until successful intravascular positioning.} An observer, blinded to the lens allocation, <...>''
    \item \texttt{tokens:} \texttt{[\textcolor{cb-burgundy}{``this'', ``was'', ``continued'', ``until'', ``successful'', ``intravascular'', ``positioning.''}, ``An'', ``observer,'', ``blinded'', ``to'', ``the'', ``lens'', ``allocation,''  <...>]}

    \item \texttt{labels: [1, 1, 1, 1, 1, 1, 1, 0, 0, 0, 0, 0, 0, 0 <...>]}
\end{itemize}

\paragraph{Performance Metric} The performance is evaluated using the macro-averaged F1-score.

\section{Papilusion}
The Papilusion pipeline involves several key steps (see \autoref{fig:pipeline}): 

\begin{enumerate}
    \item \textbf{Model Fine-Tuning:} We fine-tuned encoder models to predict the corresponding labels. Although based on the same architecture, the models were fine-tuned with different hyperparameters to create a more diverse ensemble. Each model was trained on a sequence labeling task, predicting labels for every token in the input sequence.

    
    \item \textbf{Majority Voting:} After fine-tuning multiple models, we used a majority voting strategy to aggregate their predictions. This approach involved combining the predictions from each model and selecting for each token the label that received the most votes across the ensemble of models. It is a popular ensembling method, which helps to gain better performance by using several models~\cite{wani-etal-2018-whole, k-etal-2020-ssn}. For the ensemble we chose three best models, based on the performance metric on the development dataset.
\end{enumerate}

By using encoder models fine-tuned for text classification and applying a majority voting mechanism, we aimed to improve the accuracy and robustness of our classification system.

\section{Experiments}

\paragraph{Overview} We conducted a wide series of experiments with a family of DeBERTaV3 models~\cite{he2023debertav}. We investigate experiments with bottom layers freezing~\cite{ingle-etal-2022-investigating} and the size of the input sequence as the most important hyperparameters for the efficient model selection. We used longer sequences to fit in more context for a model. Additionally, we believe that there are some dependencies between the size of the model and effectiveness of a layer freezing and using various input sequences lengths. In our experiments we conducted a detailed research and comparison of various setups.

\begin{table}[htbp]
    \centering
	\footnotesize
	\begin{tabular}{lrrr}
		\toprule
		  Model & Params & Hidden size & Layers \\
		\midrule
        Xsmall & 22M & 384 & 12\\
        Small & 44M & 768 & 6\\
        Base & 86M & 768 & 12\\
        Large & 304M & 1024 & 24\\
		
		\midrule
  
		$\mathsf{DistillBERT}$ & 66M & 768 & 6\\
		
		\bottomrule
	\end{tabular}
\caption{Model architectures comparison. The DistillBERT model was used as a baseline by the competition organizers.}
\label{table:models}
\end{table}

\begin{table*}[ht!]
	\footnotesize
    \centering
	\begin{tabular}{lrrrrrrrrrrrr}
		\toprule
		  Model size & \multicolumn{3}{c}{Xsmall} &
		\multicolumn{2}{c}{Small} & \multicolumn{3}{c}{Base} & \multicolumn{4}{c}{Large}\\
		\cmidrule(lr){2-4} \cmidrule(lr){5-6} \cmidrule(lr){7-9} \cmidrule(lr){10-13}
		Frozen layers & $\mathsf{0}$ & $\mathsf{6}$ & $\mathsf{12}$ & $\mathsf{0}$ & $\mathsf{6}$ & $\mathsf{0}$ & $\mathsf{6}$ & $\mathsf{12}$ & $\mathsf{0}$ & $\mathsf{6}$ & $\mathsf{12}$ & $\mathsf{18}$\\
		\midrule
		
		$\mathsf{256\ input}$ &  97.32 & 97.00 & \textbf{79.20} & 95.88 & 82.74 & 96.99 & 96.22 & 89.68 & - & - & 98.12 & 98.17 \\

        $\mathsf{512\ input}$ &  \textbf{98.33} & \textbf{97.71} & 78.49 & \textbf{98.81} & \textbf{83.31} & \textbf{98.55} & \textbf{98.72} & \textbf{90.66} & - & \textbf{98.83} & 98.82 & 99.07 \\

        $\mathsf{1024\ input}$ &  86.38 & 85.82 & 55.65 & 88.13 & 63.14 & 88.12 & 80.97 & 71.24 & \textbf{96.29} & - & \textbf{98.84} & \textbf{99.21} \\

        $\mathsf{2048\ input}$ &  67.87 & 62.95 & 28.99 & 65.30 & 45.47 & 66.21 & 56.40 & 50.25 & 92.87 & 92.87 & - & - \\
		
		\bottomrule
	\end{tabular}
\caption{Test results obtained after the end of competition. The symbol '-' is used to denote experiment results where with the given configuration it was not possible to achieve satisfactory or adequate outcomes.}
\label{table:test_res}
\end{table*}

\paragraph{Competition-included experiments}
During the competition we utilized competition organizers' baseline code and did several approaches of fine-tuning DeBERTa models. During the competition timeline we tried several variations of different hyperparameters such as input sequence length, model size, freezing several bottom layers of the model, learning rate scheduler. 
Our final submission brought us to 6th place with \textbf{89.83} F1 metric which was obtained with the ensemble of 3 DeBERTa-base models with different number of frozen layers and the default input sequence length (512 tokens).

\paragraph{Post-competition study}
After the end of competition we identified a tokenization issue in the code: the chunk function truncated sequence by the number of words, not tokens which resulted in information loss. We fixed our training scripts with proper tokenization in chunk function, along with some minor adjustments in the prediction function. These modifications significantly improved the models' performance, resulting in a score increase of +9.63, and enabled us to conduct a detailed study of various models and hyperparameter variations. The results are shown in the Table~\ref{table:test_res}

We used the following hyperparameters for our experiments:
\begin{itemize}
    \item Model size - \{Xsmall, Small, Base, Large\}
    \item Input sequence length - \{256, 512, 1024, 2048\}
    \item Frozen layers - \{0, 6, 12, 18\}
\end{itemize}
We fixed other hyperparameters (learning rate, scheduler, seed, validation dataset size) to perform fair comparison. In this case, we can see how changing previously mentioned  hyperparameters affects the result. Additionally, we did a series of ensemble mixtures based on majority voting technique for our best models to improve our final score.

\paragraph{Hardware specification} We run experiments on
a single GPU TESLA V100 32GB. Model fine-tuning is conducted using the transformers library~\cite{wolf-etal-2020-transformers}. The fine-tuning for DeBERTa requires approximately 4-12 hours to complete depending on the size of the model and the length of the sequence. The inference on the official test set runs within 1 hour. For DeBERTa-large model we lower the batch size down to 4, for other model versions we set the batch size to 16.

\section{Results}

\paragraph{Competition phase}
During the competition we fine-tuned base DeBERTa models with different number of frozen bottom layers and then applied majority voting for 3 models. This brought us to the 6th place (see Table \ref{table:comp_res}) and the best score among the teams below 90.0 F1 score.

\begin{table}[htbp]
    \centering
	\footnotesize
	\begin{tabular}{lr}
		\toprule
		  Place & Test score \\
		\midrule
        1st place & 99.54\\
        \multicolumn{2}{c}{...} \\
        4th place & 99.24\\
        5th place & 95.54\\
        6th place (\textbf{Our system}) & 89.83\\
        7th place & 89.82\\
        8th place & 89.67\\
		
		\midrule
  
		$\mathsf{Baseline}$ & 86.36\\
		
		\bottomrule
	\end{tabular}
\caption{Competition results.}
\label{table:comp_res}
\end{table}

\begin{table*}[htbp]
	\footnotesize
	\centering
	\begin{tabular}{llrrrrrrrrrrr}
		\toprule
		 Setup & Model & Frozen layers &
		input length & Test score \\
		\midrule
		
		a) & Xsmall & 0 & 512 & 98.33\\
        b) & Small & 0 & 512 & 98.81\\
        c) & Base & 6 & 512 & 98.72 \\
        d) & Large & 18 & 256 & 98.17\\
        e) & Large & 18 & 512 & 99.07\\
        f) & Large & 12 & 1024 & 98.84\\
        g) & Large & 18 & 1024 & \textbf{99.21}\\
        \midrule
        b) \& c) \& g) & & & & 99.28\\
        d) \& e) \& g) & & & & 99.45\\
        e) \& f) \& g) & & & & \textbf{99.46}\\
        
        \midrule

        1st place solution & & & & \textbf{99.54}\\
        2nd place solution & & & & 99.44 \\

        \midrule
  
		$\mathsf{Baseline*}$ & DistilBERT & 0 & 512 & 93.64\\
		
		\bottomrule
	\end{tabular}
\caption{Comparison of best results across various experimental configurations, denoted by letters (a, b, c, etc.). Our ensemble achieves the 2nd place compared to competition scores. '\&' sign stands for ensembling via majority voting. With '*' we mark improved baseline results obtained by ourselves after the competition.}
\label{table:main_res}
\end{table*}


		
  
		

\paragraph{Post-competition phase}

Table \ref{table:main_res} provides results of our best setups, ensembles based on those setups and comparison to the test scores on the leaderboard. More detailed results in various setups provided in the table \ref{table:test_res}, and our main outcomes are:
\begin{itemize}
    \item 1024-token sequence inputs are typically better for DeBERTa-large model; however, we found that using a 512-token sequence results in comparable quality across almost all model types.
    \item Freezing more of the bottom layers generally leads to better results for the DeBERTa-large model, while for other models it is better to not freeze the layers.
    \item Despite freezing all 12 encoder layers, we achieve a 90.66 F1 score with the Base DeBERTa. This result indicates that the task is relatively simple for these models.
    \item Ensembling the best single models leads to the score improvement.
\end{itemize}
The usage of DeBERTa models with corrected input sequence chunking allowed us to achieve significant improvements over the baseline. We got \textbf{98.33} F1 score with only Xsmall model version. It is \textbf{3 times smaller} than DistillBERT, that was used as the baseline model by competition organizers. Also we figured out that freezing of some layers could be helpful, especially for bigger models when model is already trained and have sufficient knowledge, thus we achieve our best single-model score of \textbf{99.21} with DeBERTa-large and 18/24 layers frozen. With the majority voting ensemble of our best Large models we achieve \textbf{99.46} F1 score which is compared to the second place based on the competition results. 

As it was mentioned, the smallest DeBERTa model already gets F1 score above 98, while larger ones improve just moderately depending on their settings. This suggests that the problem we are addressing might be not particularly complex, reducing the need for large models. Upon a close examination of the dataset's origin, we hypothesize that the construction process may have relied heavily on basic generation rules and may not have included sufficient filtering of simple samples. This could have resulted in notable differences among various classes within the dataset. Additionally, we believe that synonym replacement is a simple and difficult-to-control process, likely causing a significant shift in the data distribution and making the task easier. Furthermore, the task of designing and creating text that combines inputs from both humans and LMs presents a separate significant and unique challenge.


\section{Conclusion}

In this paper, we proposed Papilusion, an AI-generated text detection system designed for the DAGPap24 shared task. Our system ranked 6th in the competition, achieving an F1 score of 89.83. Post-competition enhancements, including fixing tokenization errors and optimizing model parameters, increased F1 score to 99.46. 

Through comprehensive ablation studies, we identified the most impactful hyperparameters and model configurations, leading to substantial performance improvements. 

While our experiments demonstrate that larger models achieve the best results when resources are not limited, we also found that even the small models (DeBERTa-Xsmall) exhibit promising performance metrics with minimal computational resources.




\section*{Acknowledgements}
AS's work results from a research project implemented in the Basic Research Program at the National Research University Higher School of Economics (HSE University). 
We acknowledge the computational resources of HSE University's HPC facilities.

\bibliography{custom,anthology}




\end{document}